\documentclass{article}

\usepackage{commands}
\usepackage{environments}

\usepackage[final, nonatbib]{neurips_2022}
\usepackage[numbers]{natbib}
\usepackage[utf8]{inputenc} 
\usepackage[T1]{fontenc}    
\usepackage{hyperref}       
\usepackage{url}            
\usepackage{booktabs}       
\usepackage{amsfonts}       
\usepackage{nicefrac}       
\usepackage{microtype}      
\usepackage{graphicx}

\usepackage{algpseudocode}  
\usepackage{algorithm}      
\usepackage[style=british]{csquotes}
\usepackage[inline]{enumitem} 
\usepackage{subcaption}

\makeatletter
\newcommand{\thickbar}{\mathpalette\@thickbar}
\newcommand{\@thickbar}[2]{{#1\mkern1.5mu\vbox{
  \sbox\z@{$#1\mkern-1.5mu#2\mkern-1.5mu$}%
  \sbox\tw@{$#1\overline{#2}$}%
  \dimen@=\dimexpr\ht\tw@-\ht\z@-.8\p@\relax
  \hrule\@height.8\p@ 
  \vskip\dimen@
  \box\z@}\mkern1.5mu}
}
\makeatother

\newcommand\restr[2]{{
  \left.\kern-\nulldelimiterspace 
  #1 
  \vphantom{\big|} 
  \right|_{#2} 
  }}

\title{Fully-Decentralized MADDPG with Networked Agents}

\author{
     Diego Bolliger \\
     \texttt{diegobo@student.ethz.ch} \\
     \And
     Lorenz Zauter \\
     \texttt{lzauter@student.ethz.ch} \\
     \AND
    Robert Ziegler \\
    \texttt{roziegler@student.ethz.ch} \\
}

\begin{document}

\maketitle

\begin{abstract}
    In this paper, we devise three actor-critic algorithms with decentralized training for multi-agent reinforcement learning in cooperative, adversarial, and mixed settings with continuous action spaces. To this goal, we adapt the MADDPG algorithm by applying a networked communication approach between agents. We introduce surrogate policies in order to decentralize the training while allowing for local communication during training.
    The decentralized algorithms achieve comparable results to the original MADDPG in empirical tests, while reducing computational cost. This is more pronounced with larger numbers of agents.
\end{abstract}

\section{Introduction}

Multi-agent reinforcement learning (MARL) is of significant practical value and has become a focus of research in the past years. MARL extends the ideas of Reinforcement Learning (RL) to settings where multiple agents interact in a common environment. These interactions can be competitive, where all agents work towards their own goal, cooperative, where all agents work towards the same goal, or a mixture of the two, e.g. agents working in teams against each other. This includes many practical examples, such as multi robot control, coordinated flight or control in power distribution systems.

We are interested in partially observable stochastic games (POSG), where each agent decides independently about its actions, has its own reward function and the goal is to maximize the individual reward as well as the overall reward. The reward functions are only accessible by the agents themselves and in the execution phase, there is only local information available. 
Recall that a classical single agent RL system such as a Markov decision process assumes that the agent controls all actions and has access to the entire state. Hence, the POSG setting poses two major challenges: how do agents account for missing information and how can they learn to cooperate despite being partially unaware of what the other agents are doing or aiming to do?

One actor-critic approach presented in \cite{lowe2017multi} makes use of a centralized controller which has access to all information during training, whereas the actions are deterministic and independent during execution as described above. This is achieved by adapting the deep deterministic policy gradient (DDPG) algorithm to the multi-agent setting (MADDPG).
A different approach is taken in \cite{zhang2018fully} where both learning and execution is decentralized and information is shared through a communication network. However, the authors assume that all agents observe the full state. 

Centralized training can be unfeasible in settings of large number of agents or when communication is limited. In this paper, we investigate possibilities of combining the algorithms from \cite{lowe2017multi} and \cite{zhang2018fully} with the goal of developing \emph{fully decentralized} MARL algorithms. That is, both training and  execution are decentralized.  An additional goal of this paper is to enable the algorithms to work only with local observations for each agent in training and execution, thus lifting further restrictions imposed by the previous authors.

To this end, we first devise a fully decentralized version of the MADDPG algorithm and in a second step introduce a communication network through which information can be exchanged during training. We develop these algorithms by first considering a cooperative setting, where all agents work towards a common goal. Then afterwards, we adapt them to mixed and adversarial settings. For the evaluation of all our algorithms, we conduct experiments using the multi-particle environment (MPE). \cite{terry2021pettingzoo}

In concrete terms, our contributions are as follows.
\begin{itemize}
    \item We develop and evaluate a fully decentralized version of MADDPG.
    \item We extend the fully decentralized MADDP to share critic parameters through a communication network during training. For this, we develop two different algorithms and evaluate both.
    \item Lastly, we modify all algorithms for the adversarial setting.
\end{itemize}

\section{Related Work}

In their overview paper \cite{zhang2021multiagentreinforcementlearningselective} from 2021  Zhang et. al. describe a variety of settings. Our work focuses on decentralized algorithms in the setting of stochastic games, what the authors call \enquote{networked algorithms}, because communication during training is decentralized through a (possibly) time varying network.

The MADDPG algorithm from Lowe et. al. \cite{lowe2017multi} is applicable for both cooperative and competitive settings. By deploying centralized $Q$-learning during the training phase, the algorithm is able to handle non-differentiable environment dynamics while yielding decentralized policies during execution time. This is achieved by approximating the shared $Q$ function by a neural network that has access to the policies of all agent and coordinates accordingly.
While this algorithm leads to good performance, it uses a centralized controller with its aforementioned drawbacks.
Acknowledging the centralized nature of their algorithm, the authors propose to mitigate the required access to the policies of all $N$ agents by complementing each agent with $N-1$ approximator networks which learn the policies of the other agents. This however, although being more decentralized, increases the scalability problem even further as the number of parameters scales quadratically in $N$.

In order to obtain scalability to large amounts of agents, Cui et. al. \cite{cui2024learning} maintain a distribution to model the mean field of the model. With mean field control (MFC), the agents are able to correlate their actions during training, being effectively guided by the mean field. This requires rigid assumptions on the environment. Through a radial basis function kernel, the training is then further centralized, leading effectively again to an approach that relies on centralized training to achieve decentralized execution. 

Zhang et. al \cite{zhang2018fully} take a different approach to obtain an algorithm that trains in a decentralized manner. The authors model the agents on a time varying communication network where the agents perform the actor step individually. In contrast to the other approaches presented so far, their algorithm maintains the decentralization aspect also during the training process, allowing for communication only through connected agents. More precisely, agents average their critic parameters with connected agents (i.e. locally) in a consensus step. While this approach performs well and is one of the few fully decentralized approaches to MARL, we found in our experiments that the consensus update -- effectively overriding the critics of the agents with the average of their neighbors' critics -- can lead to instability. Furthermore, Zhang et. al. require the agents to observe the full state space, thus not allowing for partially observable settings.

\clearpage
\section{Main Results}

\subsection{Problem Setting}
For our work we are interested in the general setting of partially observable stochastic games.
\begin{definition}[POSG]\label{def:posg}
    A \emph{partially observable stochastic game} (POSG) is defined as a tuple, 
    \[
        \left( \mathcal{D}, \S, \A, P, \mathcal{R}, \mathcal{O}, p_O, \delta_0\right)\,,
    \]
    where
\begin{enumerate}[label=\alph*)]
        \item $\mathcal{D} = \{ 1, \ldots, N \} = [N] $ is the set of $N\in\mathbb N$ agents.
        \item $\S$ is a finite set of states,
        \item $\A = \prod_{n\in\mathcal{D}}A^i$ is the finite set of joint actions and $A^i$ the set of actions of agent $i$,
        \item $P: \S\times\A\times\S \to [0,1]$ is the transition probability function,
        \item $\mathcal{R} = \{R^i\}_{i\in\mathcal{D}}$ is the set of individual reward functions $R^i: \S\times\A \to \R$ of the agents $ i \in [N] $,
        \item $\mathcal{O} = \prod_{i \in \mathcal{D}}O^i$ is the finite set of joint observations, where $O^i$ is the set of observations available to agent $i$,
        \item $p_O$ is the observation probability function, and 
        \item $\delta_0$ is the initial state distribution at stage $t = 0$.
\end{enumerate}
\end{definition}
During execution, as for the Dec-POMDP setting (see \cite{Oliehoek2012}) the agents of our POSG are assumed to only act based on their individual observations and to not communicate any further. We assume throughout that action spaces are \emph{continuous}, i.e. open subsets of some Euclidean space.

To facilitate cooperation between the agents in the case of our networked algorithms (algorithm~\ref{alg:decentralized_maddpg_hard} and~\ref{alg:decentralized_maddpg_soft} in~\ref{Ap:sec:madpg}) we allow communication between agents along a communication network $G_t$. Similar to \cite{zhang2018fully}, we define the networked setting below:
\begin{definition}[POSG with networked agents]\label{def:commi_network}
     A \emph{partially observable stochastic game with networked agents} is a tuple
    \[
        \left( \mathcal{D}, \S, \A, P, \mathcal{R}, \mathcal{O}, p_O, \delta_0, \{G_t\}_{t\ge 0} \right)\,.
    \]
    The first part of the tuple is a POSG, amended by a time varying undirected graph\linebreak $G_t := (\mathcal{D}, E_t) $ with the vertex set $\mathcal{D}$ and the set of edges at time $t$, $ E_t \subset \{(i,j)\in\mathcal{D}\times\mathcal{D}: i\neq j\}$.
    Communication between two agents $i,j\in\mathcal{D}$ at time $t$ is only possible if $ (i,j)\in E_t$.
\end{definition}

We note that in principle, agents could communicate through the network during action execution \emph{and} training, but we restrict ourselves to communication during training. This is without loss of generality, since the former can be included as actions and observations if needed.

\subsection{Fully-Decentralized MADDPG}
\label{maddpg_ldt}

To decentralize training of the MADDPG algorithm, see section~\ref{Ap:sec:madpg}, we propose that each agent $i$ keeps its own local replay buffer
$$\mathcal{D}^i = (o_i^j, a^j, r_i^j, o'^j_i)_{j=1}^B,$$
of size $B$, containing its own observations $o_i$, the true joint actions taken $a = (a_1, \ldots, a_N)$, its own rewards $r_i$ and next observations $o_i'$. 
The local critic $Q_i$ is then updated using solely the local replay buffer of agent $i$. 

A key problem in decentralizing training lies in how to obtain the critic targets $y_i$ with only local observations.
MADDPG needs to access the policies and observations of all agents, which is undesirable in our decentralized setting. 
The proposed policy approximation in \cite{lowe2017multi} may not need access to policies of other agents, but still requires access to all local observations while also increasing computational cost.

We propose that instead of approximating the actual policies of the other agents in the optimization target
$$ \hat y^j = r^j + \gamma Q_i^{\boldsymbol{\mu}'_i}(\textbf{x}'^j, \hat a'_1,\ldots,a_i, \ldots \hat a'_{N})|_{\hat a'_k=\boldsymbol{\hat\mu}_{i}'^k(o_k'^j)},$$
which requires knowledge of all local observation, we
let each agent learn \emph{surrogate policies} which mimic the behavior of the other agents by only using local observations of the agent in question.
Concretely, we let each agent learn a joint policy
$$
\boldsymbol{\mu}_{\theta_i}(o_i) = \left[\boldsymbol{\mu}_{\theta_i}(o_i)\right]_{k=1}^N,
$$ using the sampled policy gradient with respect to the surrogate policies
\begin{equation}\label{eqn:surrogate_gradients}
\nabla_{\theta_i} J \approx \nabla_{\theta_i}\left(\frac{1}{S} \sum_{j=1}^S  Q_i^{\boldsymbol{\mu}_i}(o^j_i, a^j)|_{a^j_k=\left[\boldsymbol{\mu}_{\theta_i}(o_i^j)\right]_k}\right).
\end{equation}
Note that the surrogate policies model the other agents as one centrally controlled actor. When interacting with the environment, each agent then uses the joint surrogate policy to select its action.
Practically speaking, we increase the output dimension of the actor network to the dimension $d_\mathcal{A}$ of the joint action space and use this output as a plug-in estimator for the joint action to create approximations for the target $y_i^j$ for the critic update, i.e.
$$
\hat{y}_i^j = r_i^j + \gamma Q_i^{\boldsymbol{\mu}'_i}(o'^j_i, a'^j)|_{a'^j_k=\left[\boldsymbol{\mu}_{\theta_i}'(o_i'^j)\right]_k}.
$$
We think of the surrogate policies as a form of imaginated, intrinsic representation of the other agents
 which help to learn a good policy as imaginary friends. This reflects intuitively human behaviour 
 in learning a team task, where it is beneficial to think about the actions of the team members to anticipate their actions. 

 Note that the critic still learns with transitions $(o_i, a, r_i, o_i')$ from the local replay buffer, where the true joint actions are used, 
 meaning that we assume that each agent's critic has access to the true joint actions. 
We remark that our algorithm, as is, can be used in the cooperative setting only, as each agent $i$ imagines the other agents' policies to maximize their own (local) reward $r_i$.

One drawback of this approach is that all agents train independently which could lead to optimizing for non-compatible policies.
We conclude that there is a trade-off between decentralization of training and cooperation.

\subsection{Fully-Decentralized MADDPG with Networked Agents}
\label{maddpg_ldtna}

The algorithm we propose in section \ref{maddpg_ldt}, in theory, may fail to lead to true cooperative behavior since each agent might optimize for a different joint policy which might contradict each other. This is problematic if the agents are supposed to cooperate. 
In the following, we propose a compromise between cooperation and decentralization in the form of networked agents, where parameters
of the critic are passed to neighbors in a possibly time varying communication network (definition~\ref{def:commi_network}). 

We represent the communication network at time $t$ by a right stochastic matrix $\boldsymbol  C_t \in \mathbb{R}^{N \times N}$, where $\boldsymbol  C_t(i,j) > 0$ if agent $i$ receives information from agent $j$ and $\boldsymbol  C_t(i,j) = 0$ otherwise.
The weights of the communication matrix $ \boldsymbol  C_t $ can be adjusted depending on the setting. If communication is not equally possible between all agents or depends on other factors (i.e. proximity), this formulation is able to model these situations as well.
Note that the network is exclusively used to pass parameters of the critic and the policy execution does not depend on the network. 

\subsubsection{Networked Agents with hard consensus update}
\label{subsubsec:hard_consensus}

One option to structure this communication, is to average the critics of neighboring agents after each learning step. Thus effectively overriding the local critics by the (weighted) average of itself and its neighbors. We call this \emph{hard consensus update}, which is performed after each learning interval. The idea is inspired by \cite{zhang2018fully}.

The consensus step is then performed, for each agent $i$, as
$$
    \mu_i \gets \sum_{j=1}^N \boldsymbol  C_t(i, j) \mu_j ,
$$
where $\mu_k$ are the network parameters of the critic of agent $k$. This means, in the case where $\boldsymbol  C_t \equiv I$, we recover fully decentralized learning as in section \ref{maddpg_ldt}. On the other hand,
 in the case where $\boldsymbol  C_t(i,j) = \frac{1}{N}, \forall i,j$, the agents effectively learn fully centralized since the same critic is used by all agents. This requires the critics to be parameterized in the same way for all agents.

 \subsubsection{Networked Agents with soft consensus update}
 \label{subsubsec:soft_consens}

 As the hard consensus update interferes quite bluntly with the learning of the individual critics, we propose a more subtle approach that leaves the integration of the communication requirement to the agents themselves. We call this \emph{soft consensus update}.

 Instead of updating the critics after every learning interval, we change the critic loss to include a penalty term based on the squared relative error of critic parameters of connected agents:
 \[
      \quad\mathcal{L}(\theta_i) = \frac{1}{S} \sum_{j=1}^S \left(y_i^j - Q_i^{\boldsymbol{\mu}}(o^j_i, a^j_i)\right)^{2} + \zeta \cdot\underbrace{\sum_{j=1}^N \boldsymbol{C}_t(i,j)\frac{\|\mu_i - \mu_j\|^2}{\|\mu_j\|^2}}_{\text{consensus penalty}} ,.
 \]
 We include a hyperparameter $\zeta$ to balance individual and collective learning. For the norm on the parameter space, we propose the standard Euclidean/Frobenius norm but other choices are possible too. For numerical stability, a small constant should be added to the the denominator.
 
\subsection{Extension to Adversarial and Mixed Settings}

As described above, the fully-decentralized algorithms can not be applied as is to adversarial and mixed settings, in contrast to the original MADDPG. This is because, when training the actor determining the surrogate actions, we maximize for a joint action by the sampled policy gradient \eqref{eqn:surrogate_gradients} that maximizes the critic value which corresponds only to the own reward of the agent.
Since we do not assume access the reward of other agents, another approach must be taken. For this, we let each agent take the assumption that adversarial agents have as goal to diminish their own reward. The actor is then updated in two steps.
For $N$ agents forming $K$ teams, where team $k$ has size $l$, consisting of agents $k_1, \ldots, k_l$ and having adversaries $\thickbar{k}_1, \ldots, \thickbar{k}_{N-l}$, the policy gradient is first sampled as 
\begin{equation}\label{eqn:surrogate_team_gradients}
\nabla_{\theta_i} J \approx \nabla_{\theta_i}\left(\frac{1}{S} \restr{\sum_{j=1}^S Q_i^{\boldsymbol{\mu}_i}(o^j_i, a^j)}{{\substack{{a^j_k=\left[\boldsymbol{\mu}_{\theta_i}(o_i^j)\right]_k}\\ {a^j_{\thickbar{k}}=\left[a^j_{\text{obs}}\right]_{\thickbar{k}}}   }}}\right),
\end{equation}

meaning that we plug in the observed adversarial actions $\left[a^j_{\text{obs}}\right]_{\thickbar{k}}$ (which we assumed to be able to observe) and only optimize according to the surrogate joint team action by taking a gradient \emph{ascent} step.

Then, directly following, we optimize for the surrogate adversarial actions, where the policy gradient is sampled according to
\begin{equation}\label{eqn:surrogate_adversarial_gradients}
\nabla_{\theta_i} J \approx \nabla_{\theta_i}\left(\frac{1}{S} \restr{\sum_{j=1}^S Q_i^{\boldsymbol{\mu}_i}(o^j_i, a^j)}{{\substack{{a^j_{\thickbar{k}}=\left[\boldsymbol{\mu}_{\theta_i}(o_i^j)\right]_{\thickbar{k}}}\\ {a^j_k
=\left[a^j_{\text{obs}}\right]_k}   }}}\right),
\end{equation}
and a gradient \emph{descent} step is taken.

\section{Empirical Results}
\subsection{Network architecture}
In order to keep our results comparable to \cite{lowe2017multi}, we implemented the algorithm with simple MLPs with five hidden layers with $256$ hidden nodes. Contrary to earlier suspicions, bigger networks or networks with a more specialized architecture such as GRUs and LSTMs did not improve results.

\subsection{The communication matrix}
\label{subsec:communication_matrix}
One of the most versatile parts of our networked algorithms is their adaptability to various communication situations. As comparability is essential, we decided to construct the communication matrix $\boldsymbol  C_t$ for $N$ agents as follows. For cooperative settings, we choose
\[
    \boldsymbol  C_t = \begin{pmatrix}
        1 - \eta & \frac{\eta}{N-1} & \ldots & \frac{\eta}{N-1} \\
        \frac{\eta}{N-1}  & \ddots &\ddots   &\vdots \\
        \vdots &\ddots & \ddots &  \frac{\eta}{N-1}\\
        \frac{\eta}{N-1}  &\ldots& \frac{\eta}{N-1} &  1 - \eta
    \end{pmatrix}
\]

where $\eta$ is a communication hyperparameter.

For mixed settings of type $1$ vs. $N$, we use the communication matrix of the form
\[
    \boldsymbol  C_t = \begin{pmatrix}
        1 & 0 & \ldots & & 0 \\
        0  & 1-\eta & \frac{\eta}{N-1} & \ldots  & \frac{\eta}{N-1}  \\
        0  & \frac{\eta}{N-1} &\ddots   &\ldots & \frac{\eta}{N-1} \\
        \vdots &\vdots & \ddots &  \ddots & \vdots\\
        0  & \frac{\eta}{N-1} &\ldots&\ldots&  1 - \eta
    \end{pmatrix}\,.
\]

\subsection{Cooperative setting}
\subsubsection{Simple Spread Environment}
\label{ssec:training_env}
For training and testing of our algorithms in a cooperative setting, we use the \enquote{\texttt{simple spread}} environment of the pettingzoo library \cite{terry2021pettingzoo}.
The environment consists of $N$ agents and $N$ landmarks on a 2D plane. The agents are controlled by the algorithms we are testing and the landmarks are stationary. The goal of the agents is to cover the landmarks while avoiding collisions.
They receive a global reward based on the minimum distance of the closest agent to each landmark and are penalized for each collision. Each agent receives the same reward
\begin{equation*}
    R = \sum_{i=1}^N \min_{j = 1, \ldots, N} \left\{ \left\| \ell_i-p_j \right\| \right\} - \sum_{i=1}^{N} \sum_{j=i+1}^{N} \indicator_{\left\{ \left\| p_i - p_j \right\| < \epsilon \right\}}\,,
\end{equation*}
where $\ell_i$ is the position of landmark $i$, and $p_j$ is the position of agent $j$, and $\epsilon$ is a collision threshold.
The observations are their relative positions to the landmarks and to the other agents, as well as their own position and velocity. The actions for each agent are to accelerate continuously in some direction and thus changing their trajectory.

\subsubsection{Simple Spread Experiment}\label{subsec:simple_spread_exp}
\begin{figure}[h]
    \centering
  \begin{subfigure}{0.5\textwidth}
    \includegraphics[width=.9\linewidth]{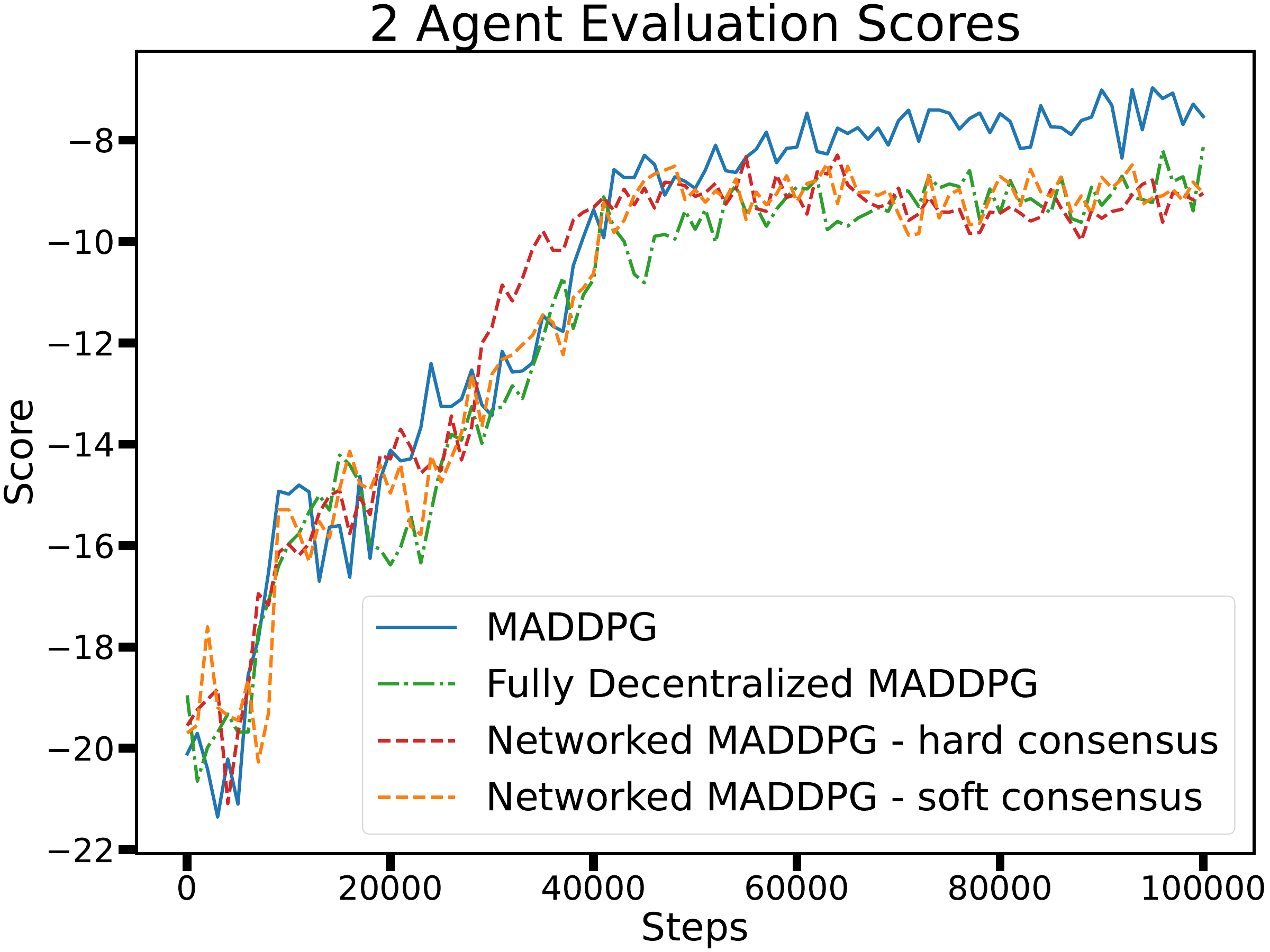}
    \caption{Two agents}
    \label{subfig:two_agents}
  \end{subfigure}%
  \begin{subfigure}{0.5\textwidth}
    \includegraphics[width=.9\linewidth]{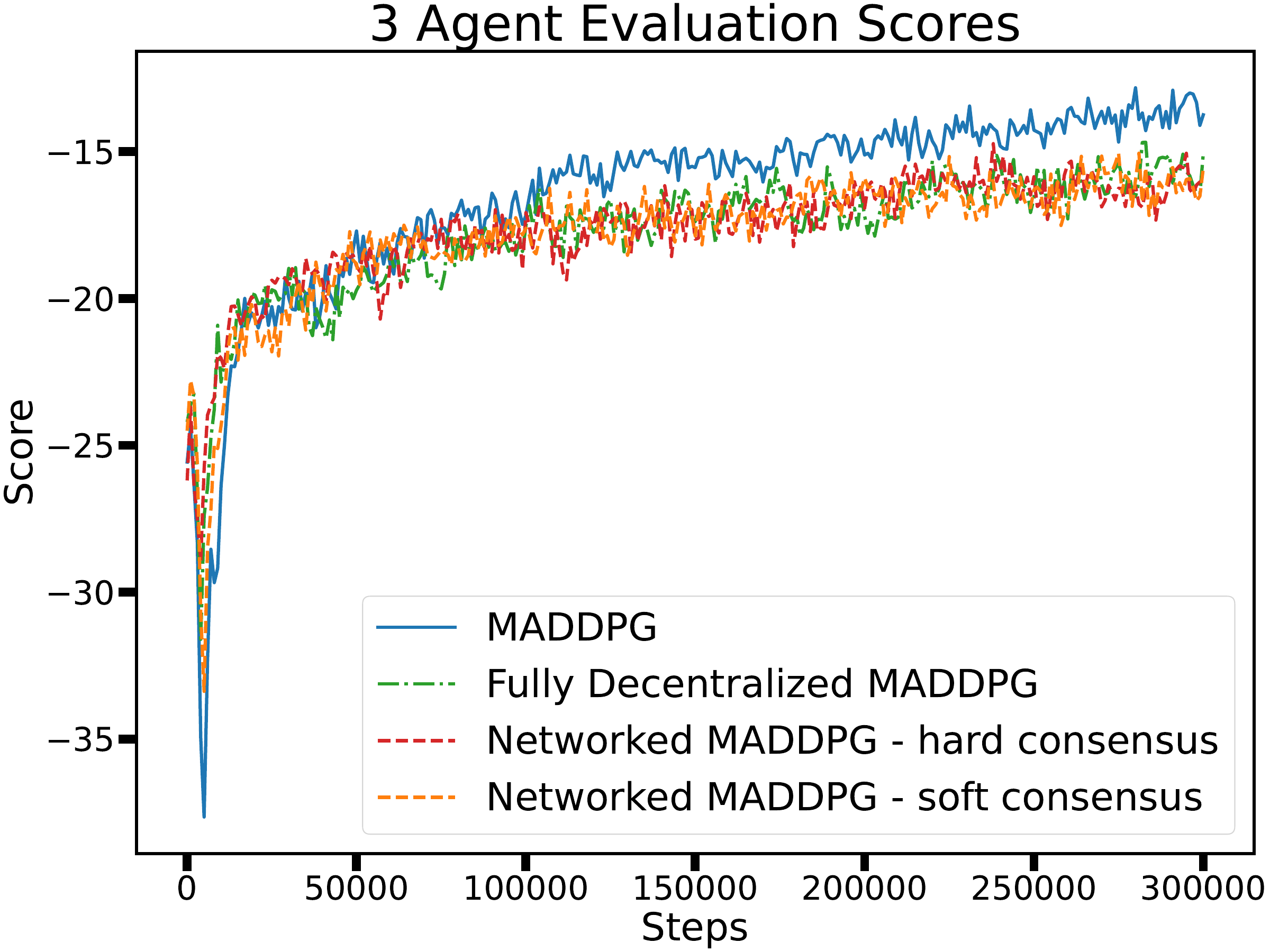}
    \caption{Three agents}
    \label{subfig:three_agents}
  \end{subfigure}

  \vspace{1ex}
  \begin{subfigure}{0.5\textwidth}
      \includegraphics[width=.9\linewidth]{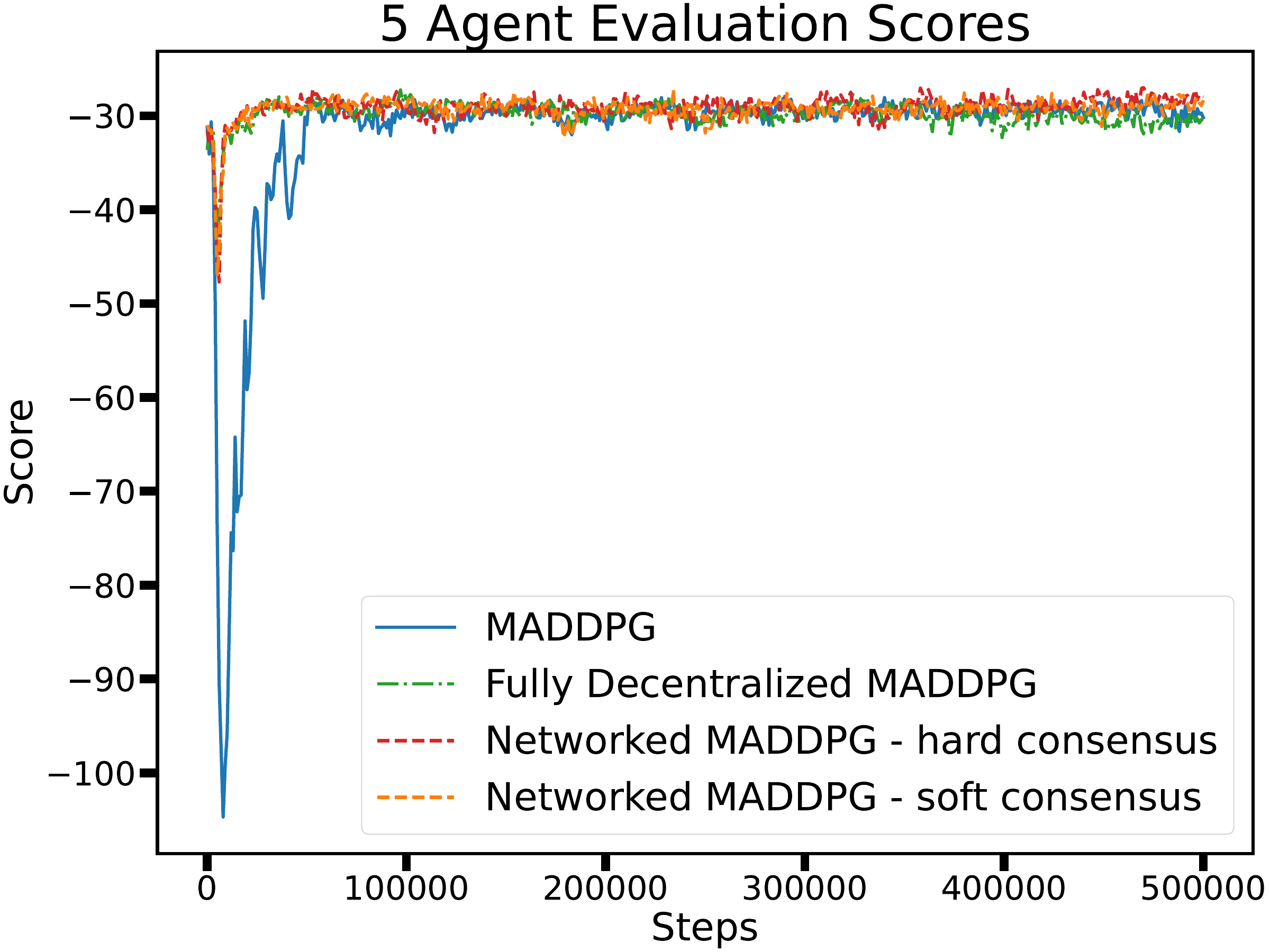}
      \caption{Five agents}
      \label{subfig:five_agents}
  \end{subfigure}%
  \begin{subfigure}{0.5\textwidth}
      \includegraphics[width=.9\linewidth]{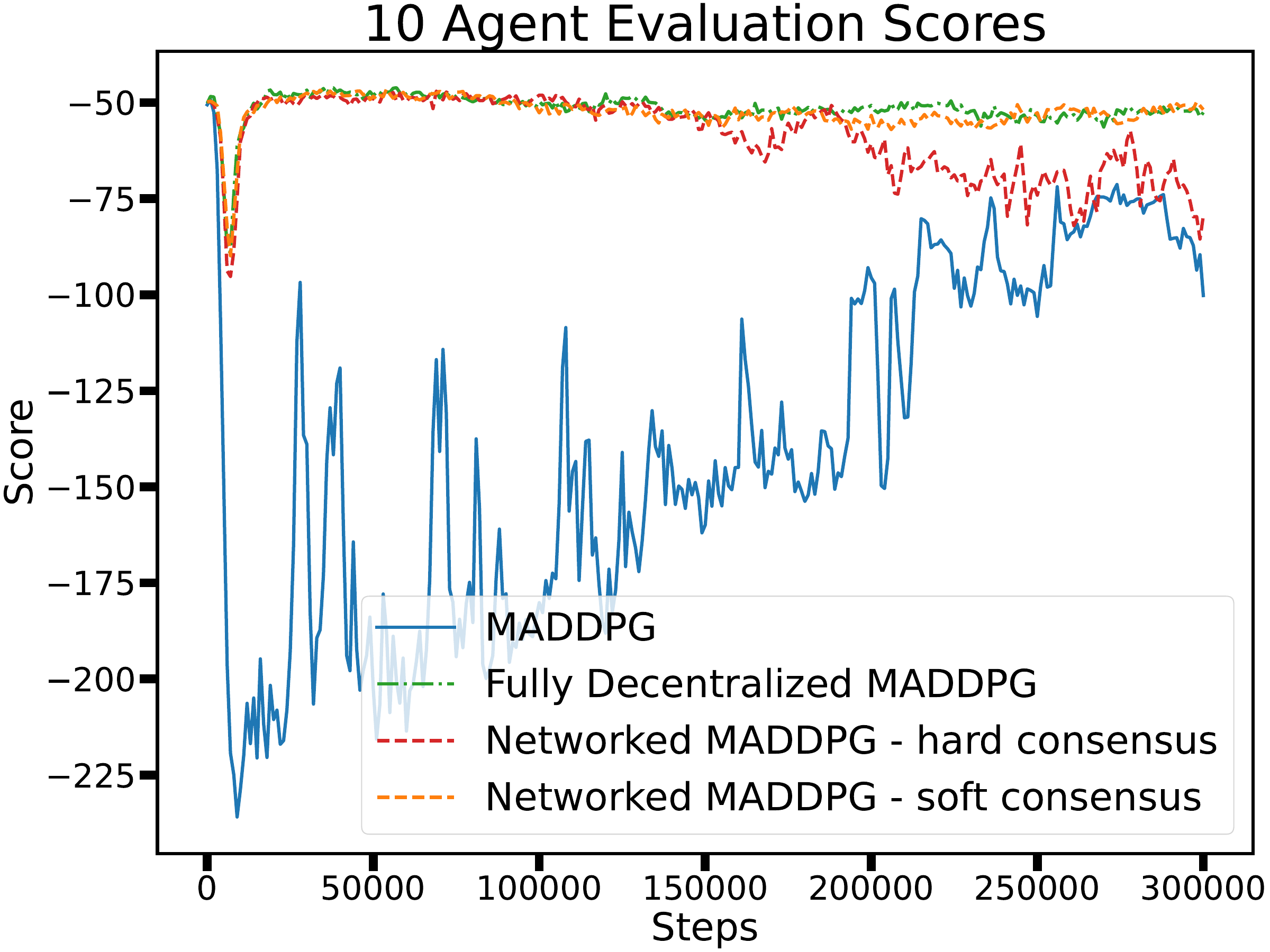}
      \caption{Ten agents}
      \label{subfig:ten_agents}
  \end{subfigure}
  \caption{Comparison of evaluation score averaged over 100 test episodes for the simple spread environment between standard MADDPG, the fully decentralized MADDPG, and the fully decentralized MADDPG with hard and soft consensus update.}
  \label{fig:score_exp}
\end{figure}

In the first set of experiments, we are interested in the obtained scores of our new algorithms in comparison with the standard MADDPG algorithm. Figure~\ref{fig:score_exp} shows how the evaluation score changes for all four algorithms when training on the experimental setup described in section \ref{ssec:training_env}. We conduct five experiments, for $ N = 2, 3, 4, 5, 10$ agents, training them over $ s = 10^5, 3\cdot 10^5, 4\cdot 10^4, 5\cdot 10^5$ and $3 \cdot 10^5$ steps respectively. The last experiment was limited due to the computational cost.

As we can see in Figure \ref{fig:score_exp} (also see Figure \ref{fig:score_exp_four_five} in the appendix), the standard MADDPG outperforms our algorithms in terms of final score for $ N = 2, 3 $ and about the same score is reached for $ N=4, 5$. However for $ N = 4, 5 $, MADDPG takes longer to converge. 

For $N=10$ agents, MADDPG and the decentralized algorithm with hard consensus update are unstable, whereas our other algorithms remain stable. However, MADDPG seems to improve towards the end and a longer training run would have been desirable (it was not possible to repeat the experiment for us due to the computational cost involved).

Since MADDPG has the most information during training and does not have to rely on guesses about the other agents behaviors, it is to be expected that the performance of MADDPG is best in the long run. However, as we see with the training runs with more than three agents, the decentralized algorithms converge much faster. We believe that this is due to the decentralization which reduces complexity of the critic by considering less information. This effect is most pronounced in the case $N=10$, where scaling becomes an issue.

The experiments with $N=10$ agents were also performed with a sparse matrix which gave similar results, see \ref{Ap:subsec:sparse}.

The fully decentralized algorithm (Algorithm \ref{alg:decentralized_maddpg} or Section ~\ref{maddpg_ldt}) performs promisingly. While converging significantly slower for fewer agents, its score tends towards the score of MADDPG. However, the networked algorithms are not able to outperform the fully decentralized algorithm. 
These train equally fast as the fully decentralized algorithm and especially for $N=5$ agents considerably faster than the (centralized) MADDPG algorithm. 

While the algorithms with hard and the soft consensus update seem to perform quite similar, the soft consensus update is more stable in case of more aggressive updating (compare~\ref{Ap:subsec:proportion_of_communication}).

However, we clearly notice diminishing returns when the number of agents increase compared to the starting scores of effectively random agents. We conclude that the MADDPG algorithm performs poorly in the simple spread already for a low number of agents. This is also confirmed by qualitative examination of the evaluation episodes, which indicate that effective policies are only learned for 2 and 3 agents.

\subsection{Adversarial and Mixed Setting}
\subsubsection{Simple Adversary Environment}
For testing adversarial and mixed settings, we use the \enquote{\texttt{simple adversary}} environment of the pettingzoo library \cite{terry2021pettingzoo}. 
There are $N$ good agents, one adversary, and $N$ landmarks from which one is a target landmark. All agents observe the relative position of landmarks and other agents. In addition, the good agents (but not the adversary) observe the relative position to the target. The adversary is rewarded by its distance to the target, and the good agents by their minimum distance. To avoid the adversary following the others to the target, the good agents need to split up to deceive the adversary. The rewards are
\begin{align*}
    R_j &=  -\min_{j = 1, \ldots, N} \left\{ \left\| \ell_\text{target}-p_j \right\|_2 \right\} - R_{\text{adv}}\\
    R_{\text{adv}} &= -\left\| \ell_{\text{target}}-p_{\text{adv}} \right\|_2,
\end{align*}
where $R_j$ are the rewards of the good agents, $p_j$ are the positions of the good agents and $\ell_\text{target}$ is the position of the target landmark, whereas $R_{\text{adv}}$ and $p_{\text{adv}}$ are the reward and the position of the adversarial.

\subsubsection{Simple Adversary Experiment}

In this set of experiments we trained the agents in a $1$ vs. $1$ and a $1$ vs. $2$ setting for $10^5$ and $3\cdot 10^5$ steps respectively. The plots are included in the appendix~\ref{Ap:subsec:adversarial_mixed}, as they show little information. Even after extensive hyper parameter exploration, the agents do not learn effective policies at all. Since this is a shortcoming of the underlying MADDPG algorithm, we cannot expect our algorithms to improve on this.

\section{Discussion and Outlook}

We developed and evaluated three fully decentralized versions of the MADDPG algorithm, which can be applied to cooperative, adversary, and mixed settings. The first algorithm, \ref{alg:decentralized_maddpg}, uses an approach of learning surrogate policies to approximate the behavior of the other agents
while not relying on further information than the own observations. The second and third algorithm, \ref{alg:decentralized_maddpg_hard} and \ref{alg:decentralized_maddpg_soft} additionally use a communication network to exchange critic parameters. The hard consensus update averages the critics of neighboring agents after each learning step, while the soft consensus update includes a consensus penalty term.

Overall, all three algorithms show promising results that are similar to the performance of the original MADDPG algorithm. 
We noticed that fully-decentralized MADDPG with networked agents and hard consensus update can be unstable if hyperparameters are not carefully chosen. This problem did not occur with the soft consensus step.
In our experiments, all algorithms are able to learn effective policies whenever the original MADDPG algorithm is able to do so.
However, we observed shortcomings of the original MADDPG algorithm in the cooperative setting for more than three agents and in the adversarial setting in general.

Even with the original MADDPG algorithm, learning a good policy is difficult and the results often lack real cooperation. By extensive experimentation, we concluded that neither the network architecture nor the hyperparameters were at fault. Therefore, 
in future work, we would like to apply the decentralization techniques we developed, i.e. learning joint surrogate policies and networked agents approaches, to other algorithms such as MAPPO, which perform considerably better in MPE environments \cite{yu2022mappo}.

\clearpage
\medskip
\bibliographystyle{plainnat}
\bibliography{ref.bib}

\clearpage
\appendix
\section{Appendix}
\subsection{MADDPG}\label{Ap:sec:madpg}
MADDPG as presented in \cite{lowe2017multi} is a version of an actor-critic algorithm for MARL with deterministic continuous policies for each agent.
\begin{theorem}[Policy gradient theorem]\label{thm:policy_gradient_thm}
    Let $ J(\theta_i) = \E_{s\sim P, a\sim\pi_{\theta_i}}[R^i] $ the expected reward of each agent $i$. Then we can write the gradient of the policy as:
    \begin{equation}\label{eq:policy_gradient}
        \nabla_{\theta_i} J(\theta_i) = \E_{s \sim \mu, a_i \sim \pi_i}\left[\nabla_{\theta_i} \log \pi_{\theta_i}(a_i|o_i) \nabla_{a_i}Q^\pi_i(\textbf{x}, a_1,\ldots, a_{N})\right]\,,
    \end{equation}
    where $ \textbf{x} = (o_1,\ldots, o_{N}) $ are the observations of all agents, $ Q_i^\pi(\textbf{x}, a_1,\ldots, a_{N}) $ is a centralized action-value function, $\pi_{\theta_i}(a_i|o_i)$ is the parameterized policy of agent $i$ utilizing only local observations and $\theta_i$ are the policy parameters for the agents.
\end{theorem}
The authors of \cite{lowe2017multi} extend this framework to deterministic 
 policies $\mu_\theta: \S \to \A$. By applying Theorem~\ref{thm:policy_gradient_thm} to deterministic policies Lowe at. al. find the actor update to be
\begin{equation}\label{eq:policy_gradient_deterministic}
    \nabla_{\theta_i} J(\boldsymbol{\mu}_{\theta_i}) = \E_{\textbf{x}, a_i \sim \mathcal{D}}\left[\nabla_{\theta_i} \boldsymbol{\mu}_{\theta_i}(a_i|o_i) \nabla_{a_i}Q^{\boldsymbol{\mu}}_i(\textbf{x}, a_1,\ldots, a_{N})|_{a_i = \boldsymbol{\mu}_{\theta_i}(o_i)}\right]\,.
\end{equation}
Further, they update the critics $ Q^\mu_i $ by minimizing the losses
\begin{equation}\label{eq:critic_loss}
    \mathcal{L}(\theta_i) = \E_{\textbf{x}, a_i \sim \mathcal{D}}\left[(y - Q^\mu_i(\textbf{x}, a_1,\ldots, a_{N}))^2\right]\,,
\end{equation}
where
\begin{equation}\label{eq:critic_target}
    y = r_i + \gamma Q^{\mu'}_i(\textbf{x}', a'_1,\ldots, a'_{N})|_{a'_k = \mu'_k(o_k)}\,
\end{equation}
where $\mathbf{\mu'} = \{\mu_{\theta_1}',\ldots,\mu_{\theta_N}'\}$ is the set of target policies with delayed parameters $\theta_i'$.
Actor, critic and target functions are all approximated with neural networks and and the gradients are estimated using samples from a replay buffer $(\mathbf{x}, \mathbf{a}, \mathbf{x}', \mathbf{a}')$ which stores past interactions between agents and environment.

\clearpage
\subsection{Algorithms: Variants of Multi-Agent Deep Deterministic Policy Gradient}
Algorithm~\ref{alg:maddpg_original} shows the original algorithm for Multi-Agent Deep Deterministic Policy Gradient (MADDPG) from \cite{lowe2017multi}.

Algorithm \ref{alg:decentralized_maddpg} is our fully-decentralized variant of the MADDPG algorithm, while algorithms \ref{alg:decentralized_maddpg_hard} and \ref{alg:decentralized_maddpg_soft} are the networked approaches with hard and soft consensus update respectively. Algorithms \ref{alg:decentralized_maddpg_mixed} to \ref{alg:decentralized_maddpg_mixed_soft} show our adaptations of the previously discussed algorithms to the mixed and adversary setting.

\begin{algorithm}
\caption{Multi-Agent Deep Deterministic Policy Gradient}\label{alg:maddpg_original}
\begin{algorithmic}[0]
    \For{episode $= 1$ to $M$}

      \State Initialize a random process $\mathcal{N}$ for action exploration
      \State Receive initial state $\textbf{x}$

      \For{$t = 1$ to max-episode-length}

        \State for each agent $ i $, select action $ a_i = \boldsymbol{\mu}_{\theta_i}(o_i) + \mathcal{N}_t $ w.r.t. the current policy and exploration 
        \State Execute actions $ a = (a_1,\ldots, a_{N}) $ and observe reward $ r $ and next state $ \textbf{x'} $
        \State Store transition $ (\textbf{x}, a, r, \textbf{x}') $ in  replay buffer $ \mathcal{D} $
        \State $\textbf{x} \gets \textbf{x}'$

        \For{agent $ i = 1 $ to $ N $}
            \State Sample a random minibatch of $ S $ samples $ \{(\textbf{x}^j, a^j, r^j, \textbf{x'}^j)\} $ from $ \mathcal{D} $
            \State Set $ y^j = r^j + \gamma Q_i^{\boldsymbol{\mu}'}(\textbf{x}'^j, a'_1,\ldots, a'_{N})|_{a'_k=\boldsymbol{\mu}_{k}'(o_k^j)}$
          \State Update critic by minimizing the loss:
                $ \mathcal{L}(\theta_i) = \frac{1}{S} \sum_{j} \left(y^j - Q_i^{\boldsymbol{\mu}}(\textbf{x}^j, a^j)\right)^{2} $
          \State Update the actor policy using the sampled policy gradient:
            \[
                 \nabla_{\theta_i} J \approx \frac{1}{S} \sum_{j} \nabla_{\theta_i} Q_i^{\boldsymbol{\mu}}(\textbf{x}^j, a^j_1,\ldots, a_i, \ldots a^j_{N})|_{a_i=\boldsymbol{\mu}_i(o_i^j)}
            \]
            \EndFor
            \State Update actor and critic target network parameters for each agent $ i $:
            \[
                \theta_i' \gets \tau \theta_i + (1 - \tau) \theta_i'
            \]
        \EndFor
    \EndFor
\end{algorithmic}
\end{algorithm}
\clearpage

\begin{algorithm}
    \caption{Fully-Decentralized MADDPG}\label{alg:decentralized_maddpg}
    \begin{algorithmic}[0]
        \For{episode $= 1$ to $M$}
          \State Initialize random process $ \mathcal{N} \in \mathbb{R}^{N} $ 
          \State Receive initial observation $ \textbf{x}:= (o_1,\ldots, o_{N}) $
          \For{$t = 1$ to max-episode-length}
              \For{agent $ i = 1 $ to $ N $}
              \State Select action $ a_i = \left[\boldsymbol{\mu}_{\theta_i}(o_i)\right]_i + \mathcal{N}_t $ w.r.t. the current policy and exploration
                  \State Execute actions $ a = (a_1,\ldots, a_{N}) $ and observe reward $ r_i $ and next observations
                        \State\qquad $ \textbf{x}' := (o'_1,\ldots, o'_{N}) $
                  \State Store $ (o_i, a_i, r_i, o'_i) $ in replay buffer $ \mathcal{D}^i $
              \EndFor
              \State $ \textbf{x} \gets \textbf{x}' $
              \If{$ t \mod \text{learning-interval} = 0$}
                  \For{agent $ i = 1 $ to $ N $}
                      \State Sample a random minibatch of $ S $ samples $ \{(o^j_i, a^j_i, r^j, o'^j_i)\} $ from $ \mathcal{D}^i $
                      \State Set $y_i^j = r_i^j + \gamma Q_i^{\boldsymbol{\mu}'_i}(o'^j_i, a'^j)|_{a'^j_k=\left[\boldsymbol{\mu}_{\theta_i}'(o_i'^j)\right]_k}.$
                      \State Update critic by minimizing the loss:
                      \[\textstyle
                          \mathcal{L}(\theta_i) = \frac{1}{S} \sum_{j=1}^S \left(y_i^j - Q_i^{\boldsymbol{\mu}}(o^j_i, a^j)\right)^{2}
                      \]
                      \State Update the actor policy using the sampled policy gradient:
                      \[\textstyle
                          \nabla_{\theta_i} J \approx \nabla_{\theta_i}\left(\frac{1}{S} \sum_{j=1}^S  Q_i^{\boldsymbol{\mu}_i}(o^j_i, a^j)|_{a^j_k=\left[\boldsymbol{\mu}_{\theta_i}(o_i^j)\right]_k}\right)
                      \]
                  \EndFor
                  \State Update actor and critic target network parameters for each agent $ i $:
                  \[
                      \theta_i' \gets \tau \theta_i + (1 - \tau) \theta_i'
                  \]
              \EndIf
         \EndFor
      \EndFor
    \end{algorithmic}
\end{algorithm}

\begin{algorithm}
    \caption{Fully-Decentralized MADDPG with Hard Consensus Update}\label{alg:decentralized_maddpg_hard}
    \begin{algorithmic}[0]
        \For{episode $= 1$ to $M$}
          \State Initialize random process $ \mathcal{N} \in \mathbb{R}^{N} $ 
          \State Receive initial observation $ \textbf{x}:= (o_1,\ldots, o_{N}) $
          \For{$t = 1$ to max-episode-length}
              \For{agent $ i = 1 $ to $ N $}
              \State Select action $ a_i = \left[\boldsymbol{\mu}_{\theta_i}(o_i)\right]_i + \mathcal{N}_t $ w.r.t. the current policy and exploration
                  \State Execute actions $ a = (a_1,\ldots, a_{N}) $ and observe reward $ r_i $ and next observations
                        \State\qquad $ \textbf{x}' := (o'_1,\ldots, o'_{N}) $
                  \State Store $ (o_i, a_i, r_i, o'_i) $ in replay buffer $ \mathcal{D}^i $
              \EndFor
              \State $ \textbf{x} \gets \textbf{x}' $
              \If{$ t \mod \text{learning-interval} = 0$}
                  \For{agent $ i = 1 $ to $ N $}
                      \State Sample a random minibatch of $ S $ samples $ \{(o^j_i, a^j_i, r^j, o'^j_i)\} $ from $ \mathcal{D}^i $
                      \State Set $y_i^j = r_i^j + \gamma Q_i^{\boldsymbol{\mu}'_i}(o'^j_i, a'^j)|_{a'^j_k=\left[\boldsymbol{\mu}_{\theta_i}'(o_i'^j)\right]_k}.$
                      \State Update critic by minimizing the loss:
                      \[\textstyle
                          \mathcal{L}(\theta_i) = \frac{1}{S} \sum_{j=1}^S \left(y_i^j - Q_i^{\boldsymbol{\mu}}(o^j_i, a^j_i)\right)^{2}
                      \]
                      \State Update the actor policy using the sampled policy gradient:
                        \[\textstyle
                          \nabla_{\theta_i} J \approx \nabla_{\theta_i}\left(\frac{1}{S} \sum_{j=1}^S  Q_i^{\boldsymbol{\mu}_i}(o^j_i, a^j)|_{a^j_k=\left[\boldsymbol{\mu}_{\theta_i}(o_i^j)\right]_k}\right)
                      \]
                  \EndFor
                  \State Update actor and critic target network parameters for each agent $ i $:
                  \[
                      \theta_i' \gets \tau \theta_i + (1 - \tau) \theta_i'
                  \]
                  \State \emph{Consensus update}:
                  \For{agent $ i = 1 $ to $ N $}
                  \State Update critic parameters:
                  \[
                      \mu_i \gets \sum_{j=1}^N \boldsymbol C_t(i, j) \mu_j
                  \]
                  \EndFor
              \EndIf
         \EndFor
      \EndFor
    \end{algorithmic}
\end{algorithm}

\begin{algorithm}
    \caption{Fully-Decentralized Networked MADDPG with Soft Consensus Update}\label{alg:decentralized_maddpg_soft}
    \begin{algorithmic}[0]
        \For{episode $= 1$ to $M$}
          \State Initialize random process $ \mathcal{N} \in \mathbb{R}^{N} $ 
          \State Receive initial observation $ \textbf{x}:= (o_1,\ldots, o_{N}) $
          \For{$t = 1$ to max-episode-length}
              \For{agent $ i = 1 $ to $ N $}
              \State Select action $ a_i = \left[\boldsymbol{\mu}_{\theta_i}(o_i)\right]_i + \mathcal{N}_t $ w.r.t. the current policy and exploration
                  \State Execute actions $ a = (a_1,\ldots, a_{N}) $ and observe reward $ r_i $ and next observations
                        \State\qquad $ \textbf{x}' := (o'_1,\ldots, o'_{N}) $
                  \State Store $ (o_i, a_i, r_i, o'_i) $ in replay buffer $ \mathcal{D}^i $
              \EndFor
              \State $ \textbf{x} \gets \textbf{x}' $
              \If{$ t \mod \text{learning-interval} = 0$}
                  \For{agent $ i = 1 $ to $ N $}
                      \State Sample a random minibatch of $ S $ samples $ \{(o^j_i, a^j_i, r^j, o'^j_i)\} $ from $ \mathcal{D}^i $
                      \State Set $y_i^j = r_i^j + \gamma Q_i^{\boldsymbol{\mu}'_i}(o'^j_i, a'^j)|_{a'^j_k=\left[\boldsymbol{\mu}_{\theta_i}'(o_i'^j)\right]_k}.$
                      \State Update critic by minimizing the loss:
                      \[
                          \quad\mathcal{L}(\theta_i) = \frac{1}{S} \sum_{j=1}^S \left(y_i^j - Q_i^{\boldsymbol{\mu}}(o^j_i, a^j_i)\right)^{2} + \sum_{j=1}^N \boldsymbol{C}_t(i,j)\|\mu_i - \mu_j\|
                      \]
                      \State Update the actor policy using the sampled policy gradient:
                      \[\textstyle
                          \nabla_{\theta_i} J \approx \nabla_{\theta_i}\left(\frac{1}{S} \sum_{j=1}^S  Q_i^{\boldsymbol{\mu}_i}(o^j_i, a^j)|_{a^j_k=\left[\boldsymbol{\mu}_{\theta_i}(o_i^j)\right]_k}\right)
                      \]
                  \EndFor
                  \State Update actor and critic target network parameters for each agent $ i $:
                  \[
                      \theta_i' \gets \tau \theta_i + (1 - \tau) \theta_i'
                  \]
              \EndIf
         \EndFor
      \EndFor
    \end{algorithmic}
\end{algorithm}

\begin{algorithm}
    \caption{Fully-Decentralized MADDPG for Mixed Settings}\label{alg:decentralized_maddpg_mixed}
    \begin{algorithmic}[0]
        \For{episode $= 1$ to $M$}
          \State Initialize random process $ \mathcal{N} \in \mathbb{R}^{N} $ 
          \State Receive initial observation $ \textbf{x}:= (o_1,\ldots, o_{N}) $
          \For{$t = 1$ to max-episode-length}
              \For{agent $ i = 1 $ to $ N $}
              \State Select action $ a_i = \left[\boldsymbol{\mu}_{\theta_i}(o_i)\right]_i + \mathcal{N}_t $ w.r.t. the current policy and exploration
                  \State Execute actions $ a = (a_1,\ldots, a_{N}) $ and observe reward $ r_i $ and next observations
                        \State\qquad $ \textbf{x}' := (o'_1,\ldots, o'_{N}) $
                  \State Store $ (o_i, a_i, r_i, o'_i) $ in replay buffer $ \mathcal{D}^i $
              \EndFor
              \State $ \textbf{x} \gets \textbf{x}' $
              \If{$ t \mod \text{learning-interval} = 0$}
                  \For{agent $ i = 1 $ to $ N $}
                      \State Sample a random minibatch of $ S $ samples $ \{(o^j_i, a^j_i, r^j, o'^j_i)\} $ from $ \mathcal{D}^i $
                      \State Set $y_i^j = r_i^j + \gamma Q_i^{\boldsymbol{\mu}'_i}(o'^j_i, a'^j)|_{a'^j_k=\left[\boldsymbol{\mu}_{\theta_i}'(o_i'^j)\right]_k}.$
                      \State Update critic by minimizing the loss:
                      \[
                          \mathcal{L}(\theta_i) = \frac{1}{S} \sum_{j=1}^S \left(y_i^j - Q_i^{\boldsymbol{\mu}}(o^j_i, a^j)\right)^{2}
                      \]
                      \State Update the actor policy for team actions (gradient ascent), using the sampled policy
                      \State\quad gradient with adversarial actions fixed:
                      \begin{equation*}
                        \nabla_{\theta_i} J \approx \nabla_{\theta_i}\left(\frac{1}{S} \restr{\sum_{j=1}^S Q_i^{\boldsymbol{\mu}_i}(o^j_i, a^j)}{{\substack{{a^j_k=\left[\boldsymbol{\mu}_{\theta_i}(o_i^j)\right]_k}\\ {a^j_{\thickbar{k}}=\left[a^j_{\text{obs}}\right]_{\thickbar{k}}}   }}}\right),
                        \end{equation*}
                        \State Update the actor policy for adversarial actions (gradient descent), using the sampled
                        \State\quad policy gradient with team actions fixed:
                        \begin{equation*}
                        \nabla_{\theta_i} J \approx \nabla_{\theta_i}\left(\frac{1}{S} \restr{\sum_{j=1}^S Q_i^{\boldsymbol{\mu}_i}(o^j_i, a^j)}{{\substack{{a^j_{\thickbar{k}}=\left[\boldsymbol{\mu}_{\theta_i}(o_i^j)\right]_{\thickbar{k}}}\\ {a^j_k
                        =\left[a^j_{\text{obs}}\right]_k}   }}}\right),
                        \end{equation*}
                  \EndFor
                  \State Update actor and critic target network parameters for each agent $ i $:
                  \[
                      \theta_i' \gets \tau \theta_i + (1 - \tau) \theta_i'
                  \]
              \EndIf
         \EndFor
      \EndFor
    \end{algorithmic}
\end{algorithm}

\begin{algorithm}
    \caption{Fully-Decentralized Networked MADDPG with Hard Consensus Update, Mixed}\label{alg:decentralized_maddpg_mixed_hard}
    \begin{algorithmic}[0]
        \For{episode $= 1$ to $M$}
          \State Initialize random process $ \mathcal{N} \in \mathbb{R}^{N} $ 
          \State Receive initial observation $ \textbf{x}:= (o_1,\ldots, o_{N}) $
          \For{$t = 1$ to max-episode-length}
              \For{agent $ i = 1 $ to $ N $}
              \State Select action $ a_i = \left[\boldsymbol{\mu}_{\theta_i}(o_i)\right]_i + \mathcal{N}_t $ w.r.t. the current policy and exploration
                  \State Execute actions $ a = (a_1,\ldots, a_{N}) $ and observe reward $ r_i $ and next observations
                        \State\qquad $ \textbf{x}' := (o'_1,\ldots, o'_{N}) $
                  \State Store $ (o_i, a_i, r_i, o'_i) $ in replay buffer $ \mathcal{D}^i $
              \EndFor
              \State $ \textbf{x} \gets \textbf{x}' $
              \If{$ t \mod \text{learning-interval} = 0$}
                  \For{agent $ i = 1 $ to $ N $}
                      \State Sample a random minibatch of $ S $ samples $ \{(o^j_i, a^j_i, r^j, o'^j_i)\} $ from $ \mathcal{D}^i $
                      \State Set $y_i^j = r_i^j + \gamma Q_i^{\boldsymbol{\mu}'_i}(o'^j_i, a'^j)|_{a'^j_k=\left[\boldsymbol{\mu}_{\theta_i}'(o_i'^j)\right]_k}.$
                      \State Update critic by minimizing the loss:
                      \[
                          \mathcal{L}(\theta_i) = \frac{1}{S} \sum_{j=1}^S \left(y_i^j - Q_i^{\boldsymbol{\mu}}(o^j_i, a^j)\right)^{2}
                      \]
                      \State Update the actor policy for team actions (gradient ascent), using the sampled policy
                      \State\quad gradient with adversarial actions fixed:
                      \begin{equation*}
                        \nabla_{\theta_i} J \approx \nabla_{\theta_i}\left(\frac{1}{S} \restr{\sum_{j=1}^S Q_i^{\boldsymbol{\mu}_i}(o^j_i, a^j)}{{\substack{{a^j_k=\left[\boldsymbol{\mu}_{\theta_i}(o_i^j)\right]_k}\\ {a^j_{\thickbar{k}}=\left[a^j_{\text{obs}}\right]_{\thickbar{k}}}   }}}\right),
                        \end{equation*}
                        \State Update the actor policy for adversarial actions (gradient descent), using the sampled
                        \State\quad policy gradient with team actions fixed:
                        \begin{equation*}
                        \nabla_{\theta_i} J \approx \nabla_{\theta_i}\left(\frac{1}{S} \restr{\sum_{j=1}^S Q_i^{\boldsymbol{\mu}_i}(o^j_i, a^j)}{{\substack{{a^j_{\thickbar{k}}=\left[\boldsymbol{\mu}_{\theta_i}(o_i^j)\right]_{\thickbar{k}}}\\ {a^j_k
                        =\left[a^j_{\text{obs}}\right]_k}   }}}\right),
                        \end{equation*}
                  \EndFor
                  \State Update actor and critic target network parameters for each agent $ i $:
                  \[
                      \theta_i' \gets \tau \theta_i + (1 - \tau) \theta_i'
                  \]
                  \State \emph{Consensus update}:
                  \For{agent $ i = 1 $ to $ N $}
                  \State Update critic parameters:
                  \[
                      \mu_i \gets \sum_{j=1}^N \boldsymbol C_t(i, j) \mu_j
                  \]
                  \EndFor
              \EndIf
         \EndFor
      \EndFor
    \end{algorithmic}
\end{algorithm}

\begin{algorithm}
    \caption{Fully-Decentralized Networked MADDPG with Soft Consensus Update, Mixed}\label{alg:decentralized_maddpg_mixed_soft}
    \begin{algorithmic}[0]
        \For{episode $= 1$ to $M$}
          \State Initialize random process $ \mathcal{N} \in \mathbb{R}^{N} $ 
          \State Receive initial observation $ \textbf{x}:= (o_1,\ldots, o_{N}) $
          \For{$t = 1$ to max-episode-length}
              \For{agent $ i = 1 $ to $ N $}
              \State Select action $ a_i = \left[\boldsymbol{\mu}_{\theta_i}(o_i)\right]_i + \mathcal{N}_t $ w.r.t. the current policy and exploration
                  \State Execute actions $ a = (a_1,\ldots, a_{N}) $ and observe reward $ r_i $ and next observations
                        \State\qquad $ \textbf{x}' := (o'_1,\ldots, o'_{N}) $
                  \State Store $ (o_i, a_i, r_i, o'_i) $ in replay buffer $ \mathcal{D}^i $
              \EndFor
              \State $ \textbf{x} \gets \textbf{x}' $
              \If{$ t \mod \text{learning-interval} = 0$}
                  \For{agent $ i = 1 $ to $ N $}
                      \State Sample a random minibatch of $ S $ samples $ \{(o^j_i, a^j_i, r^j, o'^j_i)\} $ from $ \mathcal{D}^i $
                      \State Set $y_i^j = r_i^j + \gamma Q_i^{\boldsymbol{\mu}'_i}(o'^j_i, a'^j)|_{a'^j_k=\left[\boldsymbol{\mu}_{\theta_i}'(o_i'^j)\right]_k}.$
                      \State Update critic by minimizing the loss:
                      \[
                          \quad\mathcal{L}(\theta_i) = \frac{1}{S} \sum_{j=1}^S \left(y_i^j - Q_i^{\boldsymbol{\mu}}(o^j_i, a^j_i)\right)^{2} + \sum_{j=1}^N \boldsymbol{C}_t(i,j)\|\mu_i - \mu_j\|
                      \]
                    \State Update the actor policy for team actions (gradient ascent), using the sampled policy
                      \State\quad gradient with adversarial actions fixed:
                      \begin{equation*}
                        \nabla_{\theta_i} J \approx \nabla_{\theta_i}\left(\frac{1}{S} \restr{\sum_{j=1}^S Q_i^{\boldsymbol{\mu}_i}(o^j_i, a^j)}{{\substack{{a^j_k=\left[\boldsymbol{\mu}_{\theta_i}(o_i^j)\right]_k}\\ {a^j_{\thickbar{k}}=\left[a^j_{\text{obs}}\right]_{\thickbar{k}}}   }}}\right),
                        \end{equation*}
                        \State Update the actor policy for adversarial actions (gradient descent), using the sampled
                        \State\quad policy gradient with team actions fixed:
                        \begin{equation*}
                        \nabla_{\theta_i} J \approx \nabla_{\theta_i}\left(\frac{1}{S} \restr{\sum_{j=1}^S Q_i^{\boldsymbol{\mu}_i}(o^j_i, a^j)}{{\substack{{a^j_{\thickbar{k}}=\left[\boldsymbol{\mu}_{\theta_i}(o_i^j)\right]_{\thickbar{k}}}\\ {a^j_k
                        =\left[a^j_{\text{obs}}\right]_k}   }}}\right),
                        \end{equation*}
                  \EndFor
                  \State Update actor and critic target network parameters for each agent $ i $:
                  \[
                      \theta_i' \gets \tau \theta_i + (1 - \tau) \theta_i'
                  \]
              \EndIf
         \EndFor
      \EndFor
    \end{algorithmic}
\end{algorithm}

\clearpage
\section{Further experimentation}

\subsection{Simple spread experiment for four agents}
\label{Ap:subsec:four_five}
As mentioned in section~\ref{subsec:simple_spread_exp} the evaluation scores for four agents look similar to those of five.

\begin{figure}[h]
    \centering
  \begin{subfigure}{0.5\textwidth}
    \includegraphics[width=.9\linewidth]{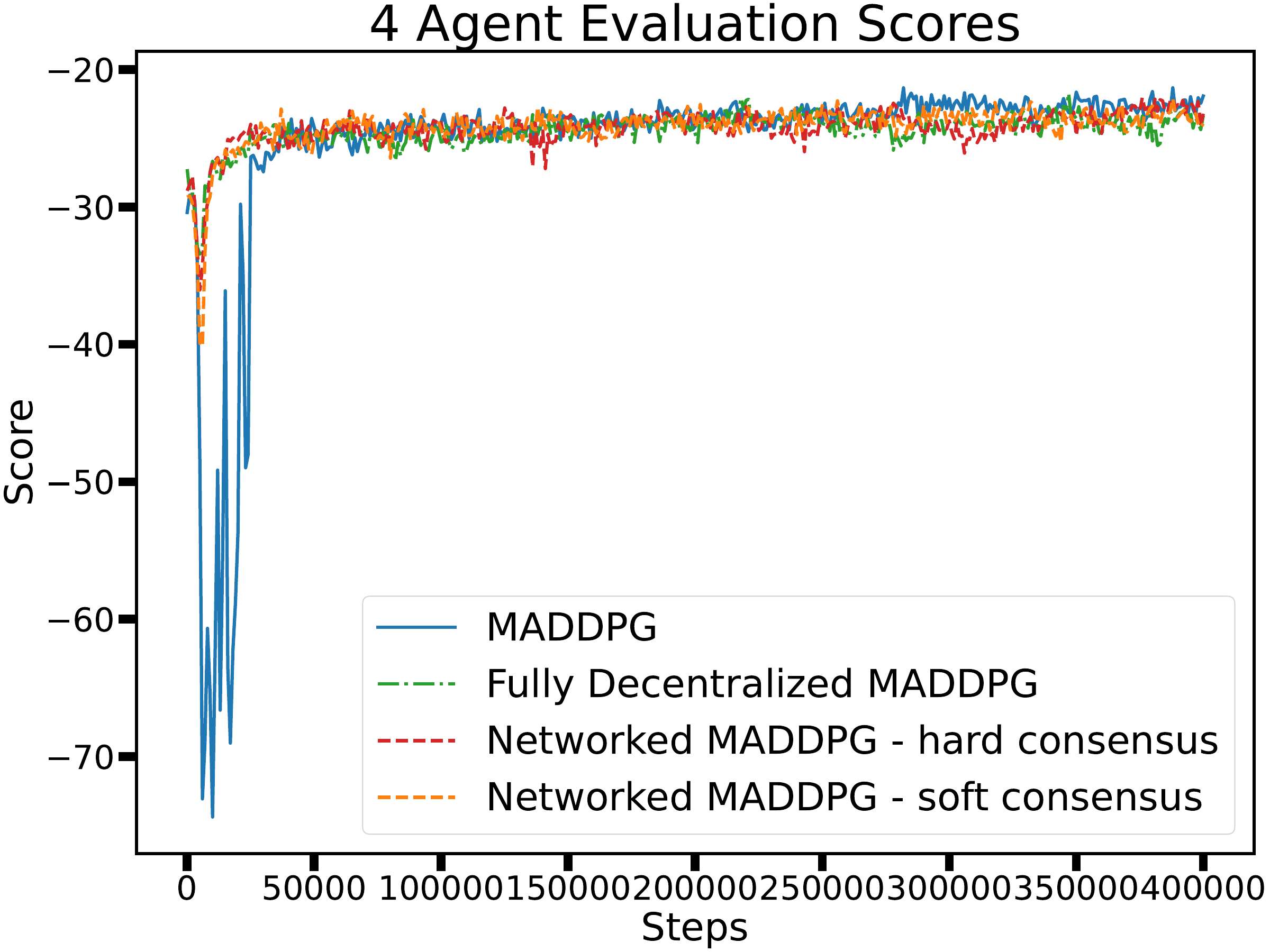}
    \caption{Four agents}
    \label{subfig:four_agents_ap}
  \end{subfigure}%
  \begin{subfigure}{0.5\textwidth}
    \includegraphics[width=.9\linewidth]{5N_Comparison_robert.pdf}
    \caption{Five agents}
    \label{subfig:five_agents_ap}
  \end{subfigure}

  \caption{Comparison of evaluation scores with four and five agents for the simple spread environment. Training was done for two agents (\subref{subfig:four_agents_ap}) over $400000$ steps and for five agents (\subref{subfig:five_agents_ap}) over $500000$.}
  \label{fig:score_exp_four_five}
\end{figure}

\subsection{Changing the communication hyperparamter}
\label{Ap:subsec:proportion_of_communication}

Altering the communication hyperparamter (see section~\ref{subsec:communication_matrix}) from $\eta = 0.001$ to $\eta = 0.05$ results in slower learning of the networked algorithms and instability of the hard consensus update at $N = 3$. For larger $N$ the hard consensus algorithm did not converge anymore while the lower rate of communication did not affect the performance of the networked algorithm with soft consensus update.

\begin{figure}[h]
    \centering
  \begin{subfigure}{0.5\textwidth}
    \includegraphics[width=.9\linewidth]{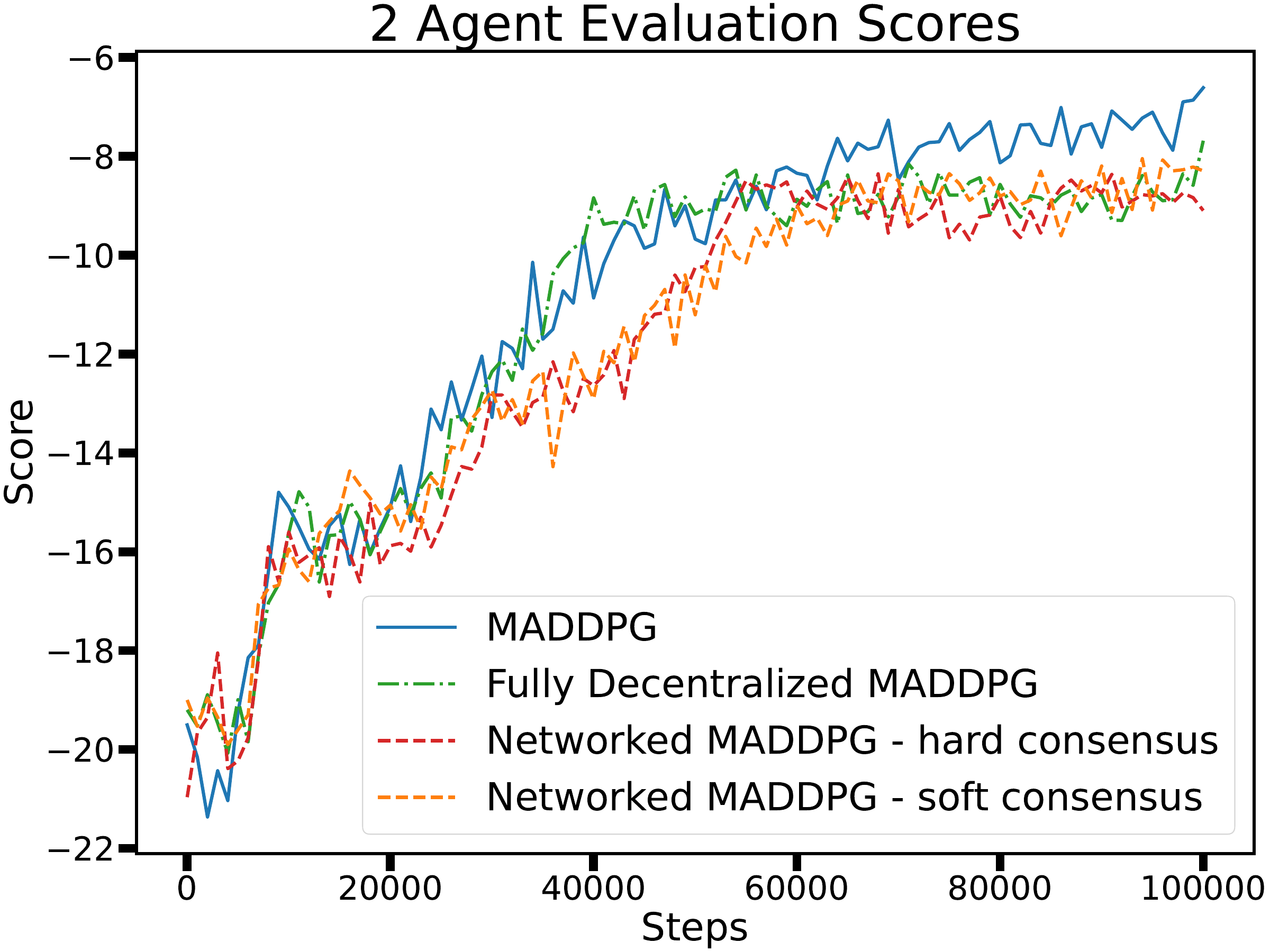}
    \caption{Two agents}
    \label{subfig:two_agents_unstable}
  \end{subfigure}%
  \begin{subfigure}{0.5\textwidth}
    \includegraphics[width=.9\linewidth]{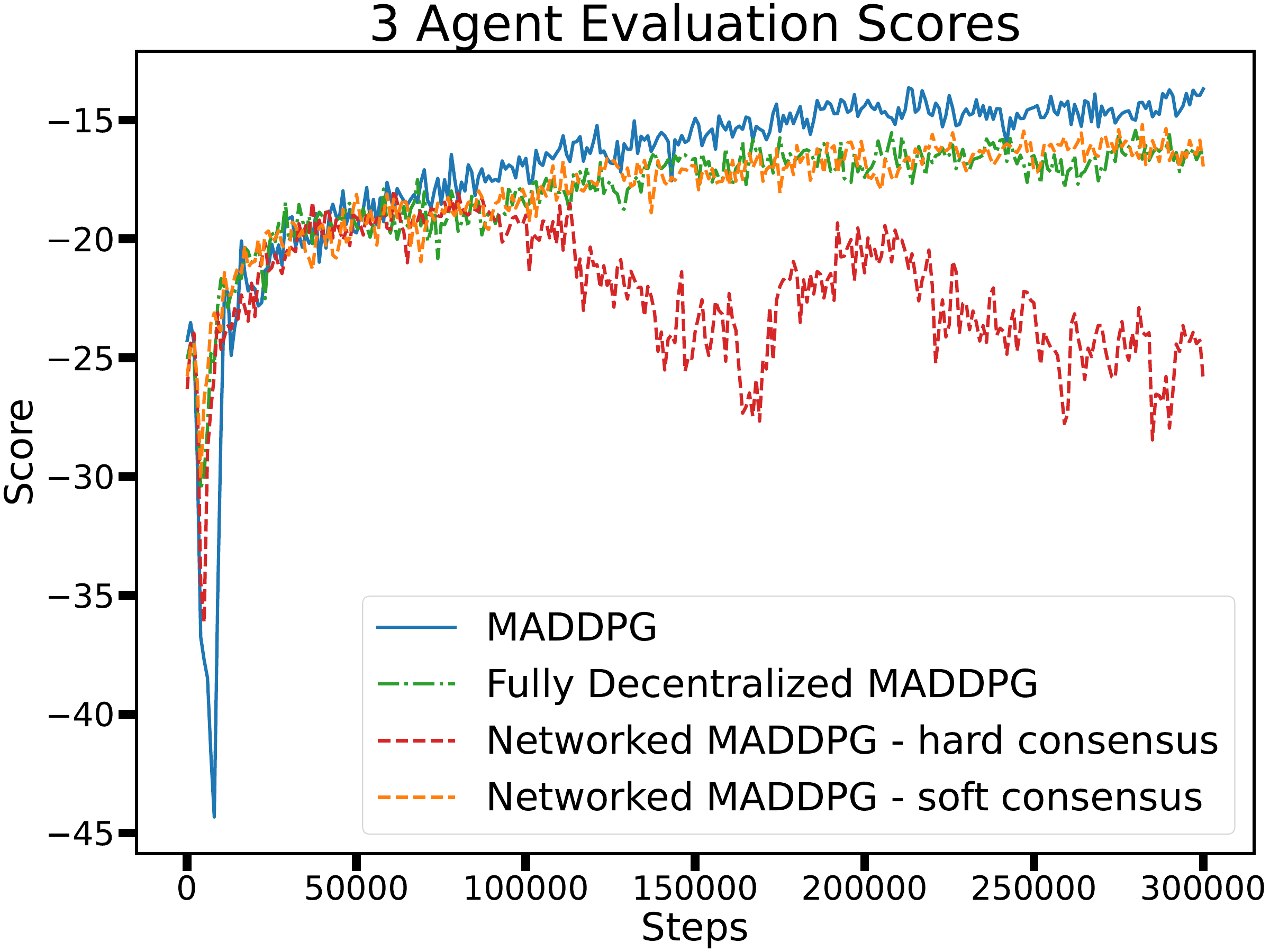}
    \caption{Three agents}
    \label{subfig:three_agents_unstable}
  \end{subfigure}

  \caption{Comparison of evaluation scores with less connectivity over training steps for the simple spread environment between standard MADDPG (blue), the fully decentralized MADDPG (green) and the fully decentralized MADDPG with hard (red) and soft (orange) consensus update. Training was done for two agents (\subref{subfig:two_agents}) over $100000$ steps and for three agents (\subref{subfig:three_agents}) over $300000$.}
  \label{fig:score_exp_unstable}
\end{figure}

\clearpage
\subsection{Sparsity of the communication matrix}
\label{Ap:subsec:sparse}
A fundamental feature of the networked algorithms is the ability to control how much agents communicate with each other during training. To test whether less communication has an influence on the training result, we compared the communication matrix described in section~\ref{subsec:communication_matrix} for the collaborative setting with a sparser matrix of the following form:
\[
    \boldsymbol  C_t = 
    \begin{pmatrix}
        1 - \eta & \eta & 0 & \ldots &0\\
        0 & \ddots & \ddots & \ddots & \vdots \\
        \vdots &\ddots &  & \ddots & 0 \\
        \vdots & & \ddots & \ddots & \eta\\
        0 & \ldots & \ldots & 0 & 1- \eta
    \end{pmatrix}\,.
\]
However, contrary to our expectations, the effect is negligible. Both for the fully connected and weighted communication network as well as for the circular and weighted network, the soft consensus update outperformed the hard consensus update. The computational cost however is lowered considerably by using the sparse communication network.

\begin{figure}[h]
    \centering
  \begin{subfigure}{0.5\textwidth}
    \includegraphics[width=.9\linewidth]{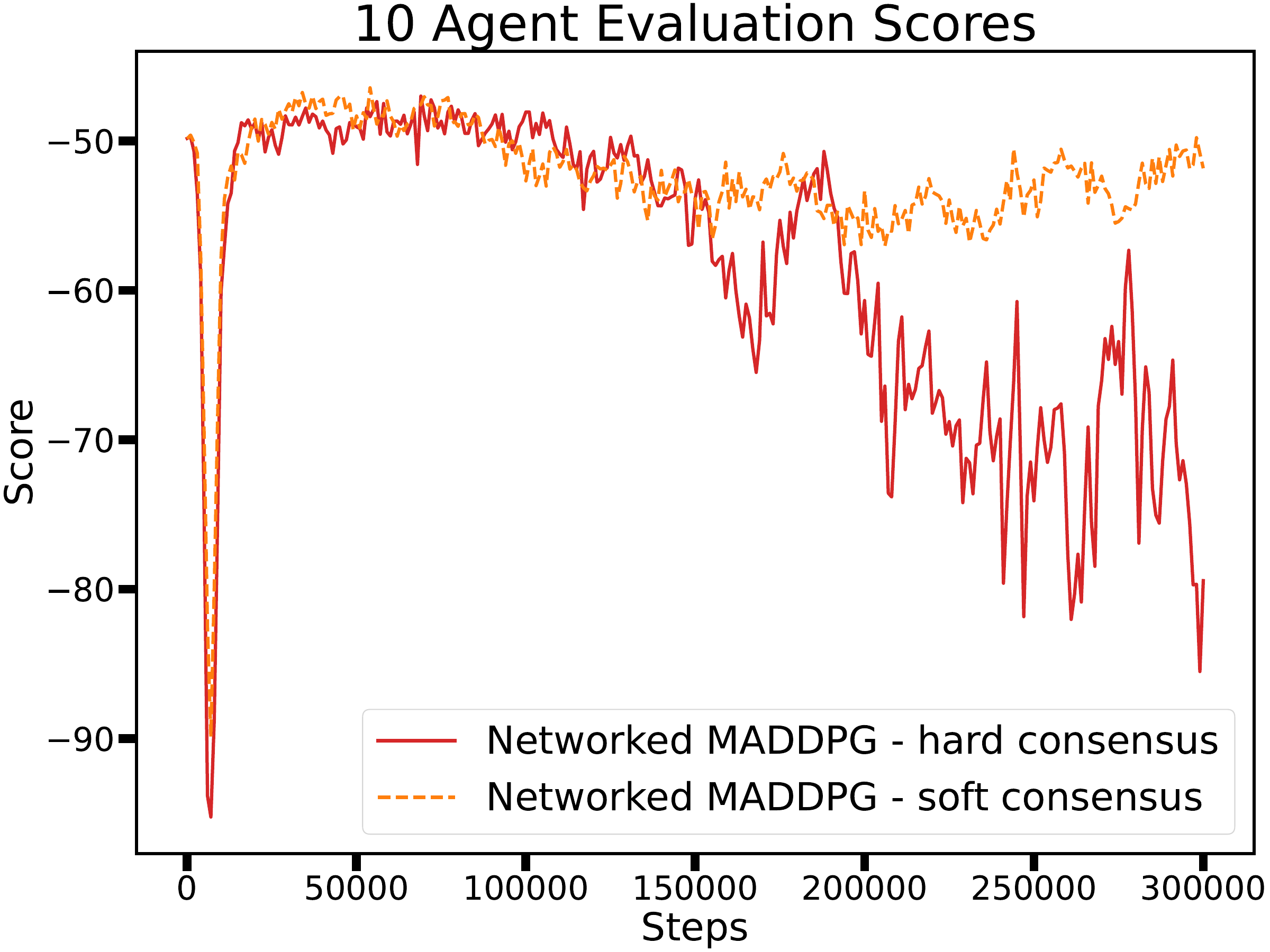}
    \caption{Fully connected communication graph}
    \label{subfig:fully_connected}
  \end{subfigure}%
  \begin{subfigure}{0.5\textwidth}
    \includegraphics[width=.9\linewidth]{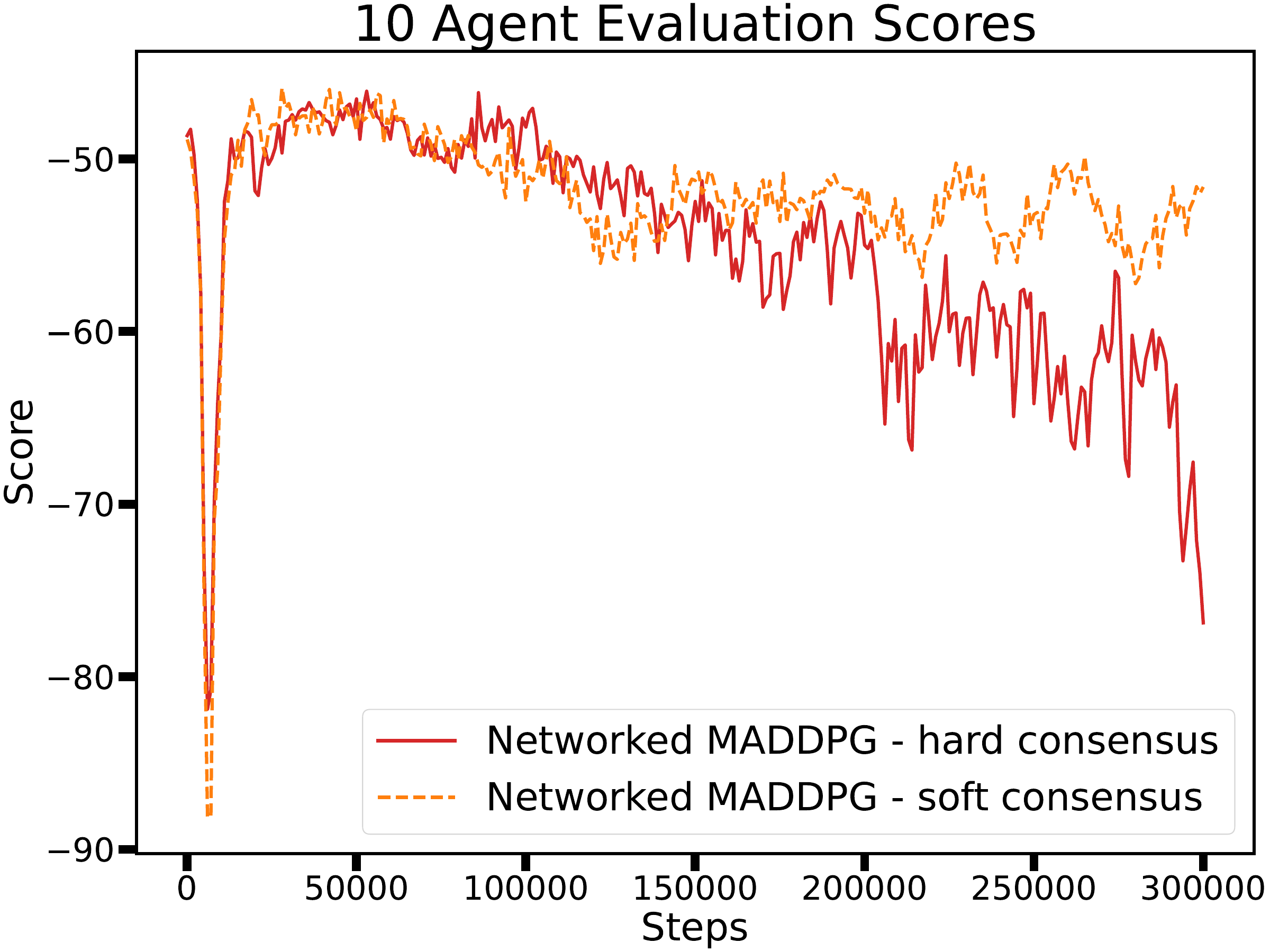}
    \caption{Sparse communication graph -- circle}
    \label{subfig:circle_graph}
  \end{subfigure}

  \caption{Comparison of evaluation scores with ten  agents for the simple spread environment with a fully connected communication graph (\ref{subfig:fully_connected}) and a circle (\ref{subfig:circle_graph}).}
  \label{fig:score_exp_four_five}
\end{figure}
\vfill

\subsection{Adversarial and mixed settings}
\label{Ap:subsec:adversarial_mixed}

\begin{figure}[b]
    \centering
    \includegraphics[width=.9\linewidth]{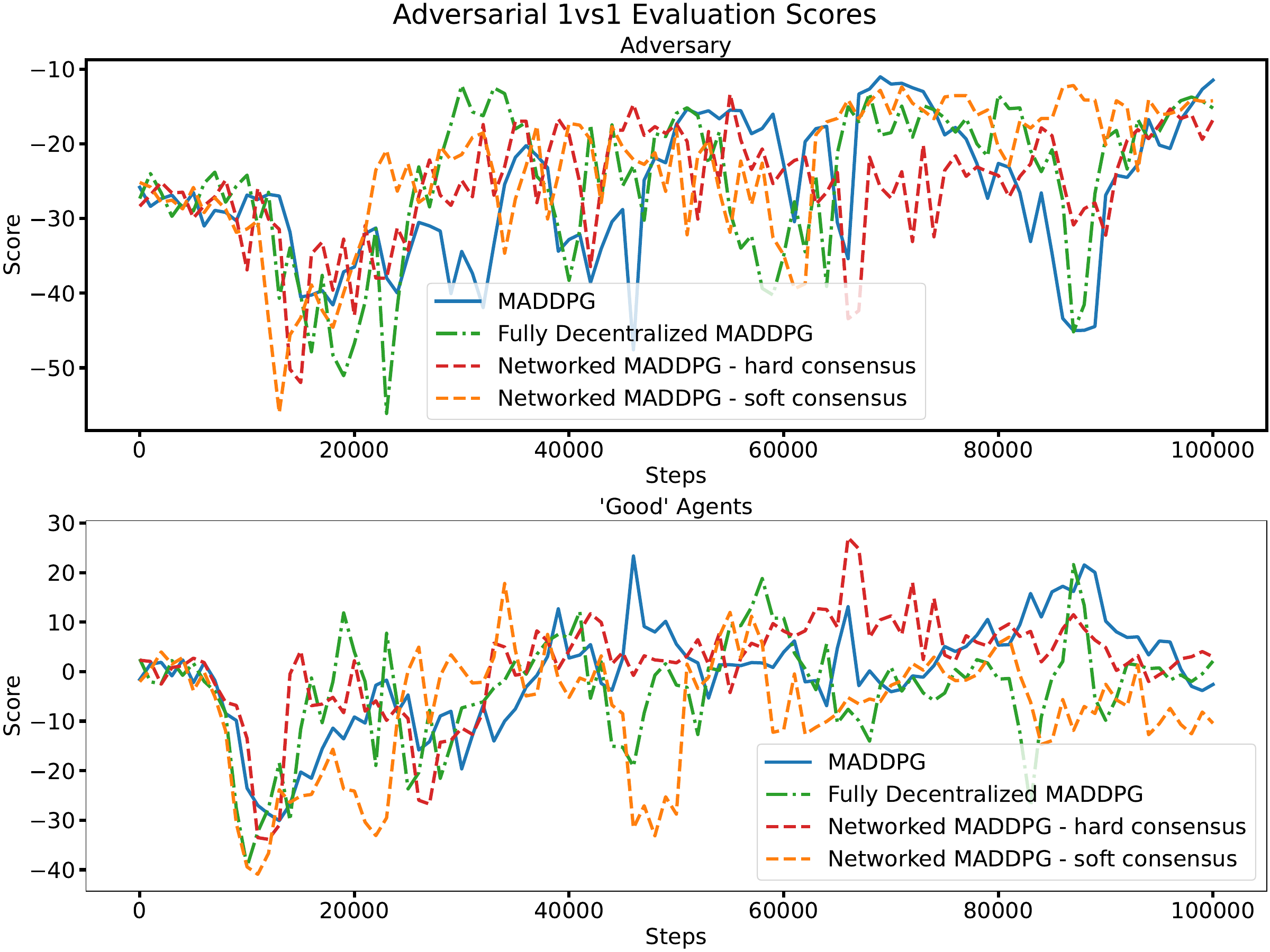}
    \caption{Performance of the algorithms in an adversarial setting of one agent against one. The two agents are shown in separate plots.}
    \label{subfig:two_agents_unstable}
\end{figure}

\begin{figure}[H]
    \centering
    \includegraphics[width=.9\linewidth]{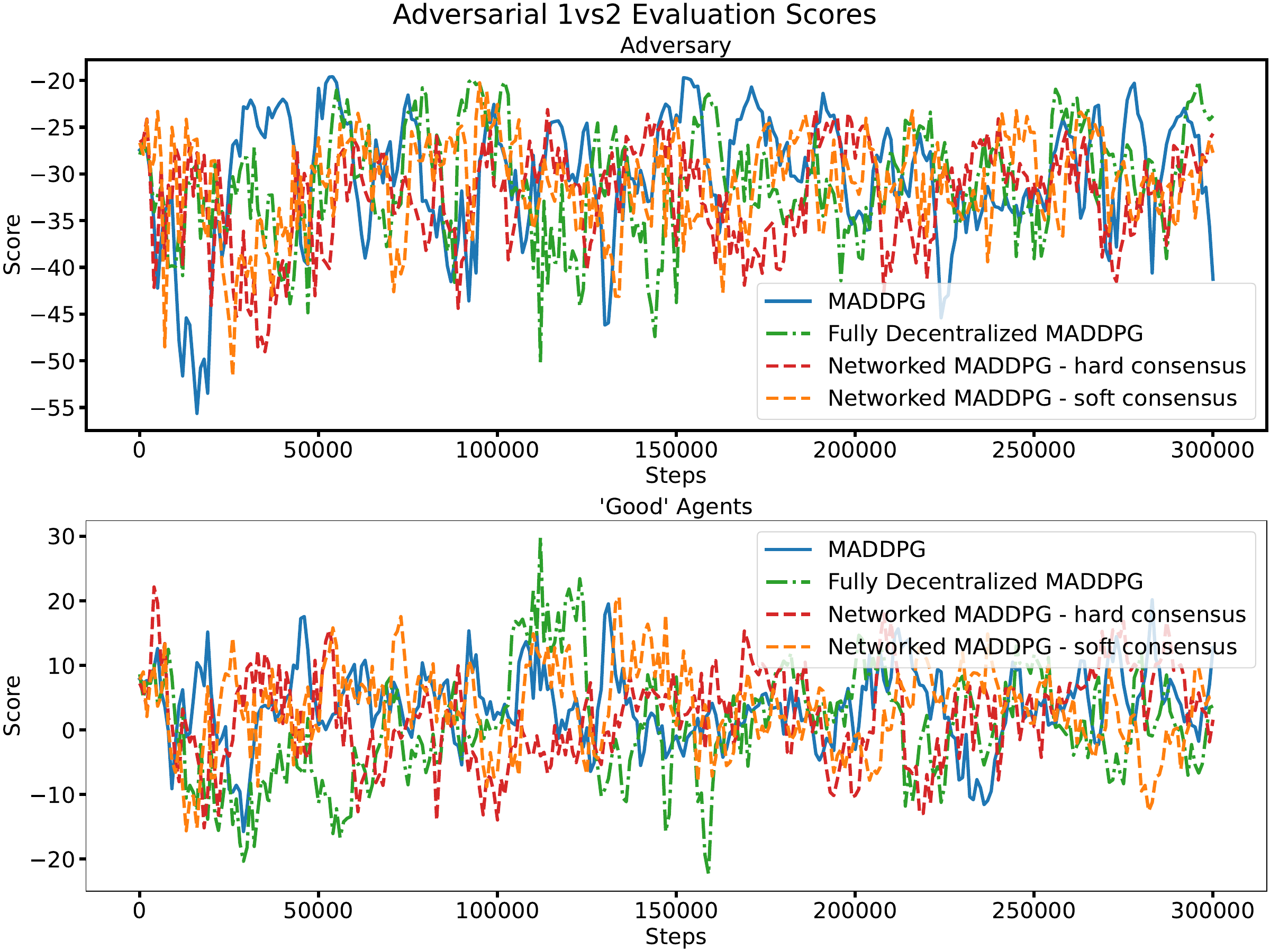}
    \caption{Performance of the algorithms in a mixed setting of one agent against two. The evaluation scores of the single agent is plotted above while the teams scores are shown below.}
    \label{subfig:three_agents_unstable}
\end{figure}

\end{document}